\documentclass{article}

\usepackage{arxiv}

\usepackage[utf8]{inputenc} 
\usepackage[T1]{fontenc}    
\usepackage{hyperref}       
\usepackage{url}            
\usepackage{booktabs}       
\usepackage{amsfonts}       
\usepackage{nicefrac}       
\usepackage{microtype}      
\usepackage{graphicx}
\graphicspath{ {./images/} }

\title{Annotating Satellite Images of Forests with Keywords from a Specialized Corpus in the Context of Change Detection}

\author{
 Nathalie Neptune \\
  IRIT, Université de Toulouse,\\
  CNRS, Toulouse INP, UT3\\
  Toulouse, France\\
  \texttt{nathalie.neptune@irit.fr} \\
   \And
 Josiane Mothe \\
  IRIT, Université de Toulouse,\\
  CNRS, Toulouse INP, UT3, INSPÉ\\
  Toulouse, France \\
  \texttt{josiane.mothe@irit.fr} \\
}

\begin{document}
\maketitle
\begin{abstract}
The Amazon rain forest is a vital ecosystem that plays a crucial role in regulating the Earth’s climate and providing habitat for countless species. Deforestation in the Amazon is a major concern as it has a significant impact on global carbon emissions and biodiversity. In this paper, we present a method for detecting deforestation in the Amazon using image pairs from Earth observation satellites. Our method leverages deep learning techniques to compare the images of the same area at different dates and identify changes in the forest cover. We also propose a visual semantic model that automatically annotates the detected changes with relevant keywords. The candidate annotation for images are extracted from scientific documents related to the Amazon region. We evaluate our approach on a dataset of Amazon image pairs and demonstrate its effectiveness in detecting deforestation and generating relevant annotations. Our method provides a useful tool for monitoring and studying the impact of deforestation in the Amazon. While we focus on environment applications of our work by using images of deforestation in the Amazon rain forest to demonstrate the effectiveness of our proposed approach, it is generic enough to be applied to other domains.
\end{abstract}


\section{Introduction}
In this paper, we propose to address the need for integrating information from satellite images and text documents for a better view of deforestation. The Sustainable Development Goals set out by the United Nations include ending deforestation \cite{UNAssembly2015}, which requires the collection and analysis of various data on the state of the Earth's forests. Earth Observation (EO) satellites provide images of forested areas, and researchers have produced numerous publications on deforestation. However, combining these sources of information can provide a better understanding of deforestation. This paper proposes generic approaches to analyze satellite images and scientific text documents to monitor changes in forests and provide solutions for environmental stakeholders and decision makers.

\begin{figure}[!htb]
\includegraphics[width=0.9\columnwidth]{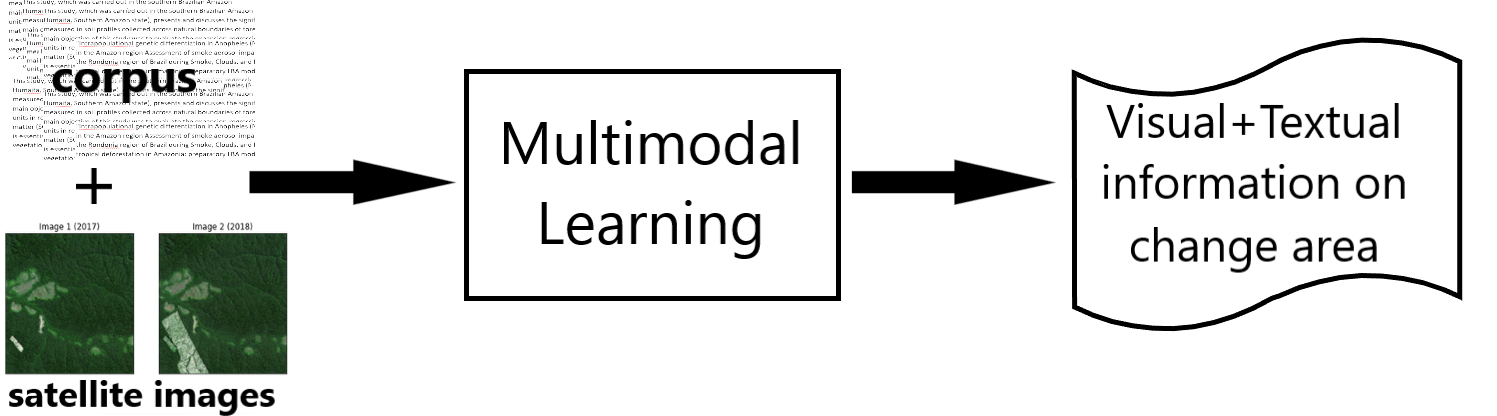}
\caption[Extracting information on forest change using text and satellite images.]{Extracting information on forest change using text and satellite images. Inputs: text corpus, satellite image pair. Outputs: change maps, textual annotations.}
\label{Fig:intro_diag_overview}
\end{figure}

Figure \ref{Fig:intro_diag_overview} illustrates the proposed approach for extracting information on forest change using text and satellite images. We focus on the use of automated methods to analyze scientific publications related to deforestation and satellite images of forested areas undergoing change. The specific characteristics of these data, such as their metadata, features, and content, make them appropriate for a variety of automatic analysis tasks. 

Our aim is to extract keywords from scientific publications related to deforestation and use them as annotations for the satellite images. We hypothesize that this combined approach will provide better insights into deforestation than using each modality separately. We aim to demonstrate that scientific publications can serve as a more resource-efficient alternative to the larger Wikipedia corpus or other web data commonly used for pre-training large language models, for the task of image annotation, in the case of deforestation. We use existing methods for joint text-image analysis and the outputs are change maps learned from the images and annotations taken from the text. We believe that this method will be useful for environmental stakeholders who need automated methods for analyzing a specific phenomenon where both images and texts can be combined as in the case of deforestation for which we have scientific publications and satellite images.

For this purpose we perform the two tasks of change detection and image annotation. Our method builds on our previous work on annotating satellite imagery for forest fires \cite{neptune2021automatic}, but introduces several improvements. For change detection, we update the encoder-decoder model from \cite{neptune2021automatic} by adding attention blocks to the decoder. As for the image annotation task, we make three main improvements. Firstly, we pre-select candidate labels using keyword extraction before predicting the annotation. Secondly, we apply our model to the phenomenon of deforestation in the Amazon rain forest, which is a different and challenging problem. Thirdly, we evaluate our method using different data sources, a different sensor (Landsat 8 vs Sentinel-2), alternative corpora, and a different word embedding model (BERT vs fastText). To achieve this, we use visual semantic embeddings (VSEs) to represent images and words in the same space, which allows for similarity comparisons. We utilize an early fusion deep neural network change detection model and image-text retrieval to find additional relevant annotations. Furthermore, we experiment with different corpora to investigate the impact of the choice of the corpus on the annotation task. As far as we know, we are the first to apply this method for change annotation in the case of deforestation in the Amazon rain forest.

We summarize our contributions as follows:
\begin{enumerate}
\item We propose combining keyword extraction with our previously proposed text retrieval approach \cite{neptune2021automatic} to annotate image pairs. This enables us to match words with a pair of images, resulting in better performance compared to annotating single images. We evaluate our approach using image pairs and demonstrate its effectiveness in our experiments.
\item We investigate the impact of using different corpora as the source of candidate annotations on our method. The selection of appropriate corpora is crucial for the annotation task, and our experiments show that different corpora have different impacts on the annotation performance.
\item We introduce novel multimodal datasets comprising satellite image pairs and scientific text to evaluate VSE models. We provide baseline performance on our deep neural network-based models using these datasets.
\end{enumerate}

\section{Annotating Satellite Images in a Change Detection Context}

\subsection{Problem Statement and Proposed Solution}
Our goal is to detect changes and annotate pairs of satellite images of the Amazon rain forest for deforestation analysis. The problem is that change detection alone may not reveal the cause and context of deforestation. Therefore, we propose a solution that combines change detection with VSE learning and information retrieval. Formally, let $\mathcal{X}_{ij}^{t_1}$ and $\mathcal{X}_{ij}^{t_2}$ be two images of $i \times j$ pixels, of the same area, taken at two different times $t_1$ and $t_2$, respectively. Let $I$ be the stacked image pair of $\mathcal{X}_{ij}^{t_1}$ and $\mathcal{X}_{ij}^{t_2}$, with dimensions $i\times j\times 2C$, where $C$ is the number of channels of each image. Our objective is: (1) to detect changes and (2) to annotate the image pair.

The first problem can be formulated as learning a function $\mathcal{F}_{\theta}$ that maps the stacked image pair $I$ to a binary change mask $M$, with dimensions $i\times j\times 1$. $\theta$ represents the set of learnable parameters of a neural network defined by the architecture of the network. The change mask $M$ indicates the locations of the pixels that have changed between the two images, with a value of 1 representing a changed pixel and 0 representing an unchanged pixel.
This problem can be cast as a binary image segmentation task, where the objective is to segment the changed pixels in the input image pair. More specifically, it is a binary change detection problem as the positive class indicates pixels that have changed from one image to the other.

For the second problem, we formulate it as a combination of VSE learning and annotation retrieval. The former involves training a supervised model to map image pairs to a high-dimensional semantic vector space. Specifically, we learn a function ${f}_{\theta}$ that maps the stacked image pair $I$ to a high-dimensional semantic vector $\mathcal{Z}_{ij} = f_{\theta}(I) \in \mathbb{R}^k$, where $k$ is the dimension of the semantic embedding. The function $f(\cdot)$ represents the learned embedding function, and $\theta$ represents the set of learnable parameters of the neural network that are updated during training to optimize the mapping from image pairs to the semantic vector space. The semantic vector $\mathcal{Z}_{ij}$ is a learned representation of the visual and semantic information contained in the image pair $I$. We use this semantic vector as the query for annotation retrieval, employing an information retrieval framework. Given a query vector $\mathcal{Z}_{ij}$, we retrieve the top $n$ relevant annotations ${w_1, w_2, \dots, w_n}$ from a pre-defined set of candidate annotations. These candidate annotations are extracted from a corpus of scientific publications related to the area of interest and change type. We either use pre-trained word embeddings or train the embeddings on the corpus to obtain a vector representation of each word. The word embeddings are then used to retrieve relevant annotations for the given query image pair. The proposed approach produces a change map and related annotations given an image pair.

\subsection{Related Work}
Change detection in image pairs is a well-studied problem in remote sensing and computer vision. The traditional approach involves comparing the two images pixel by pixel and detecting changes based on a threshold \cite{Miller1978, singh1989review}. However, this approach suffers from high false positive rates due to changes in illumination, shadows, and other environmental factors. More recently, deep learning-based methods have been proposed using convolutional neural networks (CNN) \cite{daudt2018fully, daudt2019multitask, peng2019end, de2020change, neptune2021automatic}, showing significant improvements in performance. All these methods use a variation of the U-Net architecture \cite{ronneberger2015u} due to its effectiveness in image segmentation tasks. U-Net consists of an encoder network to capture image features and a decoder network to generate a segmentation map.

Several approaches have been proposed to add or enhance the semantic information of satellite images, including the integration of ontologies into the segmentation process \cite{bouyerbou2014ontology, bouyerbou2019geographic}, and the use of geo-referenced Wikipedia articles \cite{uzkent2019learning}. However, while ontology-based approaches improved classification accuracy, they were time-consuming and required significant expert input. Similarly, the crowdsourcing approach using Wikipedia resulted in only modest improvement for semantic segmentation. VSEs have emerged as a promising solution to extract semantic information from unlabeled EO images by representing images and text in the same vector space and learning image classes based on the similarity between vector representations. A notable example of such a model is the large-scale transformer-based language model CLIP-RSICD \cite{radford2021learning,lu2017exploring} that has been pre-trained on satellite images\footnote{\url{https://github.com/arampacha/CLIP-rsicd}}. This model enables satellite images to be labeled in a zero-shot fashion or after fine-tuning. However, CLIP-like models are most effective when provided with limited candidate labels for the images; otherwise, the model is more likely to provide incorrect labels due to the large number of candidates. In our previous work \cite{neptune2021automatic} we demonstrated the effectiveness of a CNN-based VSE model in predicting labels for pairs of satellite images, in the case of wildfires. We showed that for annotating these images, higher recall of correct labels is obtained when the image pair is used as input as opposed to using a single image. In this current work, we propose combining corpus keyword extraction with our previously proposed text retrieval approach to annotate image pairs, to improve annotation results. We also investigate the impact of using different corpora as the source of candidate annotations. Additionally, we introduce novel multimodal datasets consisting of satellite image pairs and scientific text to evaluate VSE models, and provide baseline performance for our deep neural network-based models.

Our approach is suitable for a change detection dataset that is not fully annotated, meaning that some annotations may be missing or incorrect, unlike \cite{daudt2018fully, peng2019end}. We do not manually build an ontology as done in \cite{bouyerbou2014ontology, bouyerbou2019geographic}. Our approach is similar to \cite{uzkent2019learning}, but they do not perform the change detection task; instead, they classify and segment individual EO images. Our approach can be adapted to other types of remotely sensed data by modifying the network architecture as needed. Moreover, the method can be applied to various domains outside environmental sciences if image pairs with a relevant corpus of documents are available.

\subsection{Overview of our Method}
\begin{figure}[!htp]
\includegraphics[width=0.9\columnwidth]{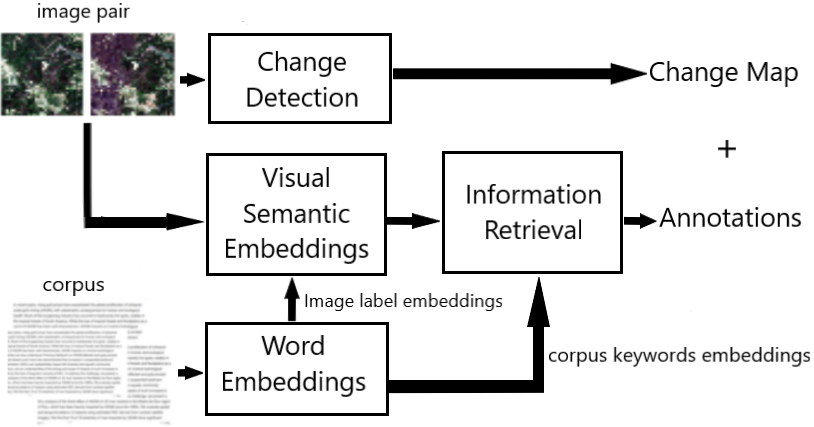}
\caption[Annotation of an image pair with keywords from scientific publications.]{Annotation of an image pair with keywords from scientific publications. An encoder-decoder model predicts a change map for a pair of images. Word Embeddings learn the vector representations of the words from the documents. VSEs learn the feature vectors of the images in the same vector space as the word embeddings. We obtain the image annotations by performing Information Retrieval; given the vector representation of the image pair as the query, we retrieve its annotations from the corpus word vectors.}
\label{Fig:overviewfig}
\end{figure}

Figure \ref{Fig:overviewfig} provides an overview of the our proposed approach. The change detection model detects the pixels that have changed between two satellite images. The VSE model learns the representation of the image pair in the same space as the embeddings of its label. The learned representation is then used for information retrieval to find the corresponding annotation from a corpus, by finding the word embeddings most similar to the image pair representation.

Our approach enables the comparison of images and words in the same embedding space, thereby facilitating the retrieval of relevant annotations through similarity measures. To achieve this, we utilize the encoding part of the early fusion deep neural network change detection model as our visual semantic model. By doing so, we can learn to represent image pairs in the word embedding space, which enables image-text retrieval. To perform image-text retrieval, we first extract the most relevant keywords from a specialized corpus chosen based on its relevance to the area and phenomenon in the images. We then compare the embeddings of these keywords with the embeddings of the images. The predicted annotations are selected as the keywords with the highest cosine similarity to the image embeddings.

\subsection{Change Detection for Image Pairs}
\label{Sec:change_detection}
Our satellite image change detection approach uses a bi-temporal method: we consider two images of the same area taken at two different times. If a change occurs between the two times, it is detected. We perform binary change detection, where the change map indicates only whether a pixel has changed or not. To achieve this, we use a U-Net architecture-based deep learning model with a ResNet encoder and attention blocks in the decoder, as proposed by \cite{daudt2018urban} and \cite{roy2018concurrent}. This model is the most simple and generic architecture that outperformed other tested models such as siamese networks, on the binary change detection task \cite{daudt2018urban,daudt2019multitask}. 

\begin{figure}[!htp]
\centering
\includegraphics[width=0.8\columnwidth]{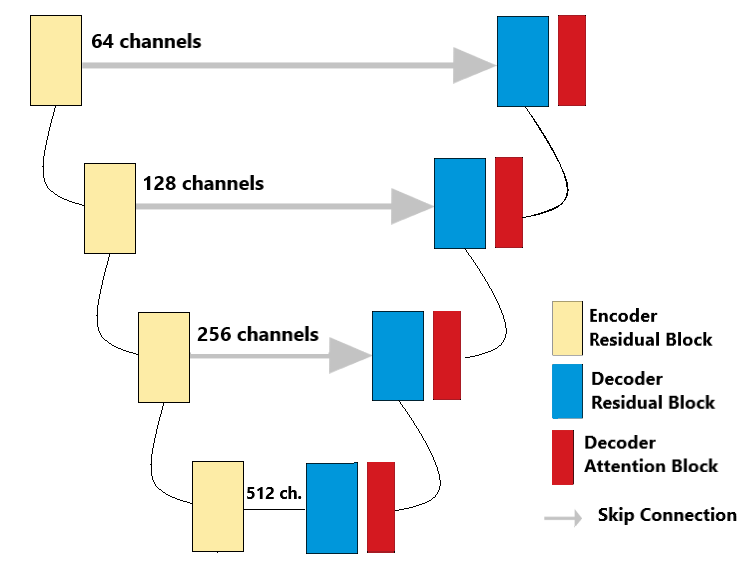}
\caption[Illustration of our encoder decoder network.]{An illustration of our encoder decoder network with residual blocks, and attention. The channels in the U-Net architecture refers to the number of feature maps in the convolutional layers. The varying number of channels from 64 to 512 represents the depth of feature maps at different stages of the U-Net.} 
\label{Fig:vse_fres_unet}
\end{figure}

The U-Net architecture is a fully convolutional neural network combining an encoder and a decoder connected with skip connections \cite{ronneberger2015u}. Different encoders can be used with the U-Net architecture, such as the Very Deep Convolutional network from the Oxford Visual Geometry Group or VGG \cite{simonyan2014very} and the Residual Network or ResNet \cite{he2016deep}. ResNet was proposed as a way to increase the depth of deep networks while improving accuracy and performance. Adding the skip connections helps avoid the degradation of performance as the network gets deeper. Figure \ref{Fig:vse_fres_unet} provides a high level overview of our encoder-decoder architecture with residual and attention blocks. Our proposed approach for change detection is similar to \cite{neptune2021automatic}, the difference is the addition of attention blocks to the decoder, which have been shown to improve the performance of fully convolutional network architectures such as U-Net for image segmentation. 

\subsection{Visual Semantic Embeddings} \label{vsedesc}
Our approach uses the encoder of the U-Net architecture along with a regression head, following a similar approach as \cite{frome2013devise}, to learn the feature vector of the image pair in the same dimension as the vector of its label. In \cite{frome2013devise}, image vector representations are projected into vectors of the same dimensions as the word vectors, and the model predicts the label vector using a similarity metric. In our case, we use the word embeddings of image labels to train the regression head of the encoder. The word embeddings are either pre-trained on large corpora namely Common Crawl, Wikipedia and other Web data, or we train them on a corpus of scientific publications selected based on their topic. We consider two specialized corpora, one related to forests in general \cite{akinyemi2018fouille} and one related to deforestation in the Amazon that we introduce in section \ref{AmzDataset}. From each specialized corpus we extract the top keywords and use them as candidate annotations for the images. When making predictions, we search for annotations among the word embeddings of the candidates. 

We employ two types of word embeddings: FastText \cite{bojanowski2017enriching} and BERT \cite{devlin2018bert}. FastText is an extension of the Word2vec model \cite{mikolov2013distributed}. Unlike Word2vec, FastText breaks words into n-grams, portions of words, and assigns each n-gram its own vector. The full word has a vector that is the sum of all its n-gram vectors. On the other hand, BERT is a transformer model that uses a self-attention mechanism. Transformers accept a sequence as input to produce an output, processing all the elements in the sequence together using self-attention. The self-attention mechanism associates each word in a sentence with every other word in the sentence, using a function of every word in the sentence, resulting in multiple embedding representations for each word based on its context.

We investigate the automatic annotation task of a pair of EO images used in change detection. We train a visual semantic model to predict the correct labels for a given image pair and use cosine similarity to measure the similarity between predicted and target annotations. We also analyze predicted annotations that were not exact matches but were among the closest word vectors.

\section{Datasets and experiments} 

\subsection{Satellite Images and Corpora}\label{AmzDataset}
\begin{figure}[!htp]
\includegraphics[width=0.8\columnwidth]{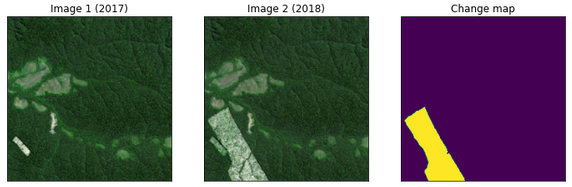}
\caption[A Patch of an Image Pair of the Amazon forest showing changes from 2017 to 2018.] {Images from the Amazon forest showing changes from 2017 to 2018. The change map, provided by \cite{de2020change}, shows the areas that have undergone change between the two years. The 2017 image already shows deforested regions. Additional deforestation can be seen in 2018 accounting for the change that can be seen in the change map.}
\label{Fig:landsatsample}
\end{figure}

The satellite images that we use show a site in the Brazilian Amazon. These images are from the Landsat 8 satellite mission and were captured by its Operational Land Imager sensor. We use the scene 230\_65 with images captured on June 21 2017, June 24 2018 and July 13 2019. The images were downloaded from the United States Geological Survey's EarthExplorer \footnote{https://earthexplorer.usgs.gov/ - EarthExplorer}. The ground truth masks were created by \cite{de2020change} using data from the Brazilian Institute of Space Research’s Project for Deforestation Mapping \cite{Shimabukuro2000}. We only use the Red, Green and Blue bands (bands 4, 3 and 2). In the ground truth, all changes from forest to another land cover type are marked as positive for change (deforestation). Figure \ref{Fig:landsatsample} shows a patch from our image dataset with the change mask. We created non-overlapping tiles of 256 $\times$ 256 pixels for each year that we used in all the experiments. We randomly selected 75\% of the tiles for training and the remaining 25\% for testing. The three years cover the same area of interest, and we selected the same tiles for all the years. This resulted in 319 tiles for training and 80 for testing, for each year.

For the corpora, we create a corpus related to our image dataset by collecting 446 publications from the Web of Science using the topic keywords "Amazon Brazil deforestation" and restricting our search to the years 2017 to 2020.  This allows us to include publications about deforestation in the Amazon that are contemporary to the images in the dataset. We refer to this corpus as the "Amazon" corpus. It does not overlap with the other large forest-related corpus \cite{akinyemi2018fouille} that we also use, which does not contain publications beyond 2016. We refer to this other corpus as the "Forest"  corpus, it contains $9722$ publications extracted from the Web of Science using the topic keyword "deforest*", spanning the years 1975 to 2016. We also indirectly use Wikipedia, Common Crawl and other Web data by using pre-trained fastText and BERT embeddings that have been trained on those datasets. The embedding dimension is 300 for fastText and 768 for BERT. The context size is 5 words for fastText and 128 tokens for BERT. 

We perform two sets of experiments with the data. The first set on change detection where we are learning to detect change on image pairs and generating the change map. The goal is to find the network that provides the best results and use its encoder in the following experiments. The second set of experiments are on learning the VSEs and finding the annotations for the image pair using the encoder from the previous experiment. 

\subsection{Experiment I : Change Detection}\label{experiment_i_cd}
The change detection task was approached as a binary image segmentation task for a pair of images for which we used a U-Net model. We trained the model with pairs of images of the same area taken at different times, with the segmentation map of the pixels as the ground truth. The two RGB images were concatenated to create a single six-channel input, and the predicted segmentation map was the model's output, where positive pixels indicated change. We tested several encoders with the U-Net architecture and chose the residual network (ResNet34) encoder based on the results. The segmentation models from PyTorch version 1.6.0 and Python version 3 implemented by Yakubovskiy\footnote{\url{https://github.com/qubvel/segmentation_models.pytorch}} were utilized to train the model with dice loss, Adam optimizer with default parameters, 200 epochs, and a 0.001 learning rate.

\begin{table}[!htp]
\centering
\begin{tabular}{lllll}
\hline
Encoder  & Precision     & Recall        & F1            & mIoU           \\ \hline
ResNet18 & \textbf{0.85} & 0.77          & 0.81          & 0.68          \\
ResNet34 & 0.83          & 0.82          & \textbf{0.83}          & \textbf{0.70} \\
ResNet50 & 0.73          & 0.76          & 0.74          & 0.59          \\
VGG11    & 0.77          & 0.86          & 0.81          & 0.68          \\
VGG16    & 0.70          & 0.84          & 0.76          & 0.62          \\
VGG19    & 0.69          & 0.83          & 0.75          & 0.60          \\ 
Daudt et al.\cite{daudt2018urban}    & 0.62          &\textbf{0.99}          & 0.76          & 0.62          \\ \hline
\end{tabular}
\caption[Precision, recall, F1, and mIoU for ResNet and VGG for change detection.]{ResNet and VGG encoders yield comparable performance measures for the binary change detection task. Resnet18 has the highest precision. Overall the smaller networks have higher precision than the larger networks. In terms of F1 and mIoU the best encoder for this dataset is ResNet34. Daudt et al. \cite{daudt2018urban} has the highest recall of all methods but the lowest precision.}
\label{Tab:change_detection_results}
\end{table}

We trained the network from scratch without any pretraining. We report the precision, recall, F1 score and mean Intersection over Union (mIoU) obtained from training the U-net model with residual network encoders (ResNet) \cite{he2016deep} and very deep convolutional networks originally from the Oxford Visual Geometry Group (VGG) \cite{simonyan2014very}. \mbox{Table~\ref{Tab:change_detection_results}} shows the results for binary change detection on the images. The values of the F1 scores varied from 0.74 to  0.83. The results are a bit varied depending on the network size, larger networks, VGG19, VGG16 and ResNet50 had F1 between 0.74 and 0.76, and mIoU between 0.59 and 0.62. While smaller networks, VGG11, ResNet18 and ResNet34 had higher F1 values between 0.81 and 0.83, and mIoU values between 0.68 and 0.70. On the basis of the results from the binary change detection task, we chose a ResNet34 encoder for the VSE learning task. 

\subsection{Experiment II : Visual Semantic Embedding}

\begin{figure*}[t!]
\centering
\includegraphics[width=0.7\textwidth]{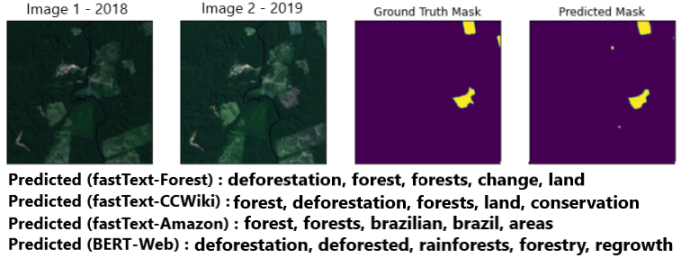}
\caption[Image pair from the Amazon dataset with the predictions made by our models.]{Sample image pair with change detection ground truth and predicted masks by the change detection model \mbox{(Section~\ref{experiment_i_cd})}, and top 5 annotations from four visual semantic models. fastText-Forest and BERT-Web correctly predict "deforestation" as top annotation, while fastText-CCWiki predicted it in second. FastText-Amazon did not predict it in the top 5, but predicted Brazil related keywords in third and fourth.} 
\label{Fig:qualitative_results_amazon}
\end{figure*}

The goal of the visual semantic task is to find a common representation for images and text in which the images and texts that are related are similar. Given an image pair, we learn the vector representation of that pair in the word vector space. We do this by performing regression with the encoder used for the binary change detection task. We add a small neural network, a regression head, to the encoder to predict a single vector for the image features. This image feature vector is of the same dimension as the word vector for the label of the image pair, which we obtain from a word embedding model. 

The regression head consists of two fully connected layers with batch normalization, dropout regularization, and ReLU activation. It takes the encoder output as input, applies adaptive max pooling, flattens it, passes it through linear layers, and outputs a vector matching the word vector size. We train the model from scratch with cosine similarity loss, Adam optimizer, 40 epochs and 0.001 learning rate.

We calculate annotation retrieval metrics to evaluate the performance of our visual semantic model as is common for visual semantic learning models \cite{frome2013devise, wang2018learning}. We report the average recall at $k$ ($R@k$) with $k$ taking values $1$, $5$, and $10$. The recall at $k$ is calculated for image-text retrieval where the image pair is the query, the result is the corresponding text. Specifically, for a given image pair, the recall at $k$ was set to $1$ if the target text was present in the top k-nearest neighbors and $0$ if not. 

We use two labels for the images, when an image pair is positive for deforestation it is labeled "deforestation" if not, it is labeled "forest". We are therefore not taking other classes that might be present into account as we do not have any other reference labels for this dataset. However, we find that this allows us to represent the majority classes of our dataset. Four variations of the visual semantic model were used in our experiments, (1) a model trained with fastText embeddings, which were trained on a the Forest corpus (fastText-Forest in Table \ref{Tab:recall_at_1_clip}), (2) a model trained with fastText embeddings, which were trained on the Amazon corpus (fastText-Amazon), (3) a model trained with fastText embeddings pre-trained on Common Crawl and Wikipedia (fastText-CCWiki), and (4) a model with BERT embeddings pre-trained on Web data (BERT-Web). Pre-trained word embedding models were not retrained on our corpora but used as is. 

Candidate annotations, for all models except fastText-Forest, were taken from the top 25 words for the Amazon corpus. The keyword extraction was done using a method similar to KeyBERT\footnote{\url{https://zenodo.org/record/4461265}} where the words for which the embeddings are most similar to the embeddings of the abstract of the scientific publications are selected. For fastText-Forest, the candidate annotations were taken from the Forest corpus in the same manner. We used the top keywords as candidate annotations to limit the possible annotations and the possibility of erroneous predictions, while still keeping the most relevant candidates. This keyword extraction step allows us to find, for each corpus the top words that are more likely to be relevant to our images while limiting the noise for our annotation task. Table~\ref{tab:word_list} shows the first 10 out of the top 25 keywords for the Amazon and Forest corpora. We do not apply any post-processing to the keyword list which is why we can see similar keywords. 

\begin{table}[htbp]
    \centering
    \begin{tabular}{l|l|l|l}
        \hline
        \multicolumn{2}{c|}{\textbf{Amazon Corpus}} & \multicolumn{2}{c}{\textbf{Forest Corpus}} \\
        \hline
        deforestation & brazilian      & forest        & forests \\
        forest        & environmental  & land          & changes \\
        amazon        & change         & deforestation & carbon \\
        brazil        & species        & change        & data \\
        land          & conservation   & species       & areas \\
        \hline
    \end{tabular}
    \caption{Top keywords extracted from two corpora.}
    \label{tab:word_list}
\end{table}

We report annotation retrieval metrics. Table \ref{Tab:recall_at_k_amz_top25} shows the results obtained with fastText and BERT embeddings, with visual semantic models trained from scratch. For recall at 1, the highest value of 0.70 is obtained with the fastText-Forest and fastText-CCWiki models. For recall at 5, the BERT-Web model reaches the highest value of 0.99. For recall at 10 the fastText-Amazon model reaches the highest value of 1.

\begin{table}[!htp]
\centering
\begin{tabular}{llrrr}
\hline
\begin{tabular}[c]{@{}l@{}}Word \\ Embedding \\ Model\end{tabular} & \begin{tabular}[c]{@{}l@{}}Corpus for \\Training the \\ Word Embeddings\end{tabular}                        & R@1  & R@5  & R@10 \\ \hline
fastText                                                  & Forest                                                                                   & 0.70 & 0.95 & 0.96 \\

fastText                                                  & Amazon                                                                                   & 0.68 & 0.68 & 1.00 \\

fastText                                                  & CC. + Wiki.                   & 0.70 & 0.94 & 0.94 \\

BERT                                                      & \begin{tabular}[c]{@{}l@{}}Web Data\end{tabular} & 0.65 & 0.99 & 0.99 \\ \hline
\end{tabular}
\caption[Recall at 1, 5 and 10 for the visual semantic models predicting annotations for the images.]{Recall results for visual semantic models using different word embeddings, evaluated at 1, 5, and 10. FastText embeddings outperform BERT at recall at 1, while BERT performs best at recall at 5. The model using fastText embedings trained on a small corpus (fastText-Amazon) performs worse than the other models at recall at 5.}
\label{Tab:recall_at_k_amz_top25}
\end{table}

We can evaluate our models qualitatively by looking at a sample of the obtained results. In Figure~\ref{Fig:qualitative_results_amazon} we see the change detection map obtained by the U-Net-ResNet34 change detection model along with the annotations predicted by the variations of the visual semantic model (ResNet34 with regression head). Only the fastText-Forest and BERT-Web models predicted the true annotation correctly as the first annotation. 

We compare our models to CLIP-RSICD \cite{radford2021learning, lu2017exploring} by testing on the case where the candidate annotations are only the two labels present in the image dataset that we are using for the tests, namely "deforestation" and "forest". We report the recall at 1, for this case we do not report recall at 5 or 10 because we restricted the candidate annotations to the two labels. Table \ref{Tab:recall_at_1_clip} shows the recall at 1 for all the models tested. CLIP-RSICD reaches a recall at 1 value of 0.49 compared to our models that reach values from 0.65 to 0.75.

\begin{table}[!htp]
\centering
\begin{tabular}{ll}
\hline
\begin{tabular}[c]{@{}l@{}}Model Name\end{tabular} & R@1  \\ \hline
fastText-Forest                                      & 0.71 \\
fastText-Amazon                                      & 0.68 \\
fastText-CCWiki                                      & 0.75 \\
BERT-Web                                             & 0.65 \\
CLIP-RSICD*                                          & 0.49 \\ \hline
\end{tabular}
\caption[When only using the two labels as candidate annotations, the fastText model trained on Common Crawl and Wikipedia improved its performance from 0.70 to 0.75 in recall at 1 compared to when it used the top 25 words of the Amazon corpus as the candidate annotations.]{Improved performance of fastText-CCWiki for annotation prediction with only 2 candidate annotations. BERT-Web and the fastText-Amazon corpus show same performance, while fastText-Forest slightly improves. CLIP-RSICD underperforms our models (*trained on single images and tested on post-change image).}
\label{Tab:recall_at_1_clip}
\end{table}

\section{Conclusion}
In this paper, we presented a method for predicting relevant annotations for pairs of satellite images undergoing visible changes. We use deep neural networks for feature extraction and annotation. We use word vectors to predict relevant keywords, either by training embeddings on the given corpus or using pre-trained embeddings. However, the performance of our method is sensitive to the corpora on which the word embedding model is trained. Our proposed method can be applied to larger and more diverse satellite image datasets, allowing for a variety of changes. One limitation of our work is the relatively small size of the image dataset, which may hinder the ability of the models to generalize to different images from different sensors. In future work, we could improve the predicted annotations with post-processing to handle inflected word forms and synonyms, also we could apply reranking to get the most relevant annotations. Furthermore, for the specific case of deforestation, benchmark datasets containing multi-temporal images are lacking, and their availability would be crucial for further advancements in automatic change detection methods for forests.

\section*{Acknowledgments} 
We thank the reviewers for their thoughtful reviews, Bissmella Bahaduri for helping with the images, and the AI4AGRI project (Horizon Europe, Grant ID: 101079136) for its support.

\bibliographystyle{unsrt}  


\end{document}